# Image Denoising Using Interquartile Range Filter with Local Averaging

Firas Ajil Jassim

*Abstract*—Image denoising is one of the fundamental problems in image processing. In this paper, a novel approach to suppress noise from the image is conducted by applying the interquartile range (IQR) which is one of the statistical methods used to detect outlier effect from a dataset. A window of size k×k was implemented to support IQR filter. Each pixel outside the IQR range of the k×k window is treated as noisy pixel. The estimation of the noisy pixels was obtained by local averaging. The essential advantage of applying IQR filter is to preserve edge sharpness better of the original image. A variety of test images have been used to support the proposed filter and PSNR was calculated and compared with median filter. The experimental results on standard test images demonstrate this filter is simpler and better performing than median filter.

## I. INTRODUCTION

Image quality improvement has been a concern throughout the field of image processing. Images are affected by various types of noise [1]. Image noise may be defined as any corrosion in the image signal, caused by external disturbance. Thus, one of the most important areas of image restoration is that cleaning an image spoiled by noise. The goal of suppressing noise is to discard noisy pixels while preserving the soundness of edge and information of the original image. Understanding the characteristics of noise helps in determining the pattern of noise appears in an image [15]. Therefore, a variety of image filtering methods have been proposed [5][3][6][17][2][14][9]. Noise filtering can be viewed as replacing every noisy pixel in the image with a new value depending on the neighboring region. The filtering algorithm varies from one algorithm to another by the approximation accuracy for the noisy pixel from its surrounding pixels [8].

The proposed algorithm in this paper focuses on how to effectively detect the salt and pepper noise and efficiently restore the image. The mechanism adopted by the proposed scheme consists of determining whether a pixel is noise or not based on some predefined threshold and calculated values. Once pixels are detected as noise in previous phase, their new value will be estimated and set in noise reduction phase.

## II. IMAGE DENOISING

Image denoising is the process of finding unusual values in digital image, which may be the result of errors made by external effects in image capturing process. Many text books in image processing include chapters about image noise and enhancement [10][12][19]. Actually, identifying these noisy values is an essential part of image enhancement. In the past three decades, a variety of denoising methods have been proposed in the image processing. In spite of these methods are very different, but they tried to remove the noisy pixels without affecting the edges, as much as possible, [13]. One of the most common filters is the median filter [11][8]. Median filter is very effective in removing salt and pepper and impulse noise while preserving image details. Median filter is performed as replacing a pixel with the median value of the selected neighborhood. In particular, the median filter performs well at filtering outlier points while leaving edges unharmed [13]. One of the undesirable properties of the median filter is that it does not provide sufficient smoothing of nonimpulsive noise [7]. Also, when increasing window size this may imply to blur edges and details in an image [18].

## III. INTERQUARTILE RANGE IQR

The Five Number Summary is a method for summarizing a distribution of data [20]. The five numbers are the minimum, the first quartile $Q_1$, the median, the third quartile $Q_3$, and the maximum. A box and whisker plot will clearly show a five number summary [4]. The IQR is the range of the middle 50% of a distribution. It is calculated as the difference between the upper quartile and lower quartile of a distribution. Since an outlier is an observation which deviates so much from the other observations. Therefore, any outliers in the distribution must be on the ends of the distribution, the range as a measure of dispersion can be strongly influenced by outliers. One solution to this problem is to eliminate the ends of the distribution and measure the range of scores in the middle. Thus, the IQR will eliminate the bottom 25% and top 25% of the distribution, and then measure the distance between the extremes of the middle 50% of the distribution that remains. IQR is a robust measure of variability [4]. The general formulas for calculating both $Q_1$ and $Q_3$ are given as:

$$Q_1 = \frac{n+1}{4} th \ ordered \ observation \qquad (1)$$

$$Q_3 = \frac{3(n+1)}{4} th \ ordered \ observation \qquad (2)$$

## IV. PROPOSED IQR FILTER

In this article, a novel filter based on the concept of the Interquartile range which is one of the measures of dispersion used in statistics that calculates variation between elements of a data set. In order to apply IQR filter, a window of size k×k was used to implement the proposed method. First, the pixels in the k×k window are sorted in ascending order in order to calculate the first and third quartiles, $Q_1$ and $Q_3$ respectively [20]. Second, the IQR is calculated by subtracting $Q_1$ from $Q_3$. Third, all the pixels that lie outside the IQR are treated as suspected pixels (SP). Those suspected pixels may be pass through a permission procedure to check weather they are noisy or not. This could be shown in the next section.

**Firas A. Jassim**, Management Information Systems Department, Irbid National University, Irbid, Jordan.

# Image Denoising Using Interquartile Range Filter with Local

## A. Permission Procedure

Actually, not all the pixels outside the IQR are noisy image. A threshold may be established to permit the external pixels (the pixels outside the IQR) to be in or out. The permission procedure is implemented in two sides which are left and right, i.e. $Q_1$ and $Q_3$. According to left side, the difference between $Q_1$ and the suspected pixel is calculated. If $|Q_1\text{-SP}|<T_1$, then the pixel is not noisy, otherwise it is. On the other hand, the same procedure is repeated for the right hand with $Q_3$. Therefore, two thresholds ($T_1$ and $T_2$) may be found to determine the truly noisy pixels. As an example, an arbitrary 8×8 window size from a random image was chosen to apply the previously mentioned procedure, table (1).

TABLE 1 ARBITRARY 8×8 WINDOW SIZE FROM A RANDOM IMAGE

| 103 | 103 | 102 | 100 | 99 | 99 | 103 | 0 |
|---|---|---|---|---|---|---|---|
| 103 | 255 | 103 | 102 | 101 | 101 | 103 | 105 |
| 102 | 103 | 105 | 105 | 103 | 102 | 255 | 104 |
| 101 | 104 | 106 | 107 | 106 | 104 | 103 | 103 |
| 100 | 103 | 107 | 108 | 106 | 104 | 103 | 102 |
| 100 | 103 | 106 | 107 | 105 | 103 | 102 | 102 |
| 100 | 102 | 105 | 0 | 103 | 102 | 102 | 102 |
| 100 | 102 | 104 | 103 | 102 | 101 | 101 | 102 |

The first quartile was found to be ($Q_1$=102) and the third quartile was ($Q_3$=104). Hence, IQR=104-102=2. Now, after transform the 8×8 block into a vector of size 64 and sorting it, the suspected pixels corresponding to the left side are 0, 0, 99, 99, 100, 100, 100, 100, 100, 101, 101, 101, 101, and 101 because they are less than $Q_1$ and hence outside IQR from left. Obviously, 99, 100 and 101 are not highly differing from $Q_1$; therefore, they are not noisy pixels and must be inside. Mathematically speaking, |102-99|=3, |100-102|=2, and |102-101|=1 which are all have small difference with $Q_1$. So, if a threshold $T_1$ was determined such that the difference of the suspected pixels is less than $T_1$. Also, all pixels higher than $T_1$, i.e. the two 0's, since |102-0|=102>$T_1$. As a result, the noisy pixels from the left side are (0,0). The same procedure could be applied to the right side and getting (255,255) as right noisy pixels, figure (1).

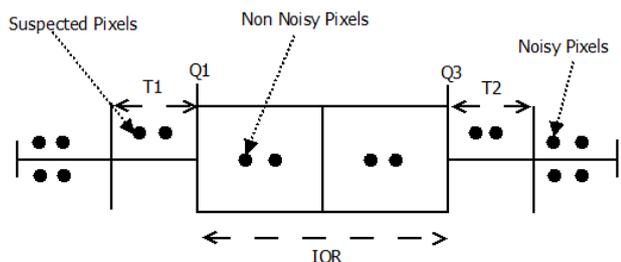

Figure 1 IQR with T1 and T2

## B. Estimating Noisy Pixelss

After the determination of the noisy pixels, the estimation method used to donate a value for each noisy image is the local averaging [16]. First, the noisy image could be classified into three types. According to figure (2), the three noise types are: corner noise (A, C, G, and I), border noise (B, D, F, and H) and interior noise (E). For the corner noise pixels, the estimation could be obtained by summing all the surrounding values (which are always three) and dividing them by 3. While for the border noise, the surrounding pixels are 5. Hence, the average for each surrounding pixels could be found. Finally, the interior noise pixels are surrounded by nine points. As an example, the estimation of the corner noise pixel (0), upper right, in figure (1), is computed as summing all the surrounding three pixels (103+103+105)/3=103.67≈104 which is a very sophisticated value.

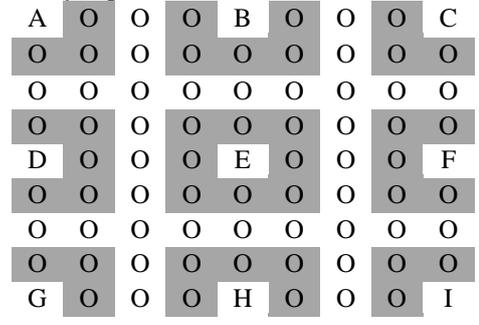

Figure 2 Three noise types

The noisy image may be represented as:

$$x_{ij} = \begin{cases} x_{corner}, & x_{ij} \in [A,C,G,I] \\ x_{borber}, & \in [B,D,F,H] \\ x_{interior}, & x \in [E] \\ x_{original}, & otherwise \end{cases} \quad (3)$$

The estimation of the noisy pixels could be obtained using local averaging as:

$$y_{ij} = \begin{cases} avg(3\ surrounding\ pixels\ of\ x_{corner}) \\ avg(5\ surrounding\ pixels\ of\ x_{border}) \\ avg(8\ surrounding\ pixels\ of\ x_{interior}) \end{cases} \quad (4)$$

## C. Noisy Neighbors Problem

Since the noise imposed randomly, the noise pixels may be neighbors in the image array. Therefore, the procedure of local averaging could be risky because of including another noisy pixel in the summation which is wrong. Hence, some procedure to get rid of the noisy neighbor just during the local averaging is very important. According to figure (3), both A and B are noisy pixels. As mentioned previously, the local averaging is used to estimate the value of the noisy pixel A by finding the local averaging of the surrounding pixels to A which are 84, B, 85, 87, and 86. But B is also a noisy image and this will affect the average directly. As an example, if the value of A is 0, then (84+0+85+87+86)/5 ≈ 52 which is very far from the nearest neighbors. So, by neglecting B and calculating the summation for all the surrounding pixels without B as (84+85+87+86)/4 ≈ 86 and that is seems to be rational approximation.

| 84 | A | 86 |
|---|---|---|
| B | 85 | 87 |
| 84 | 84 | 86 |

Figure 3 Noisy Neighbors

According to equation (3), the estimation of the noisy pixels could be reformulated as:

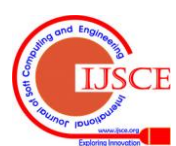

$$y_{ij} = \begin{cases} avg(3\,surrounding\ pixels\ of\ x_{corner}), if\ pixel \neq noise \\ avg(5\,surrounding\ pixels\ of\ x_{border}), if\ pixel \neq noise \\ avg(8\,surrounding\ pixels\ of\ x_{interior}), if\ pixel \neq noise \end{cases} \quad (5)$$

### D. IQR Algorithm

For each window of size k×k do the following:
1. Compute $Q_1$, $Q_3$, and IQR distance
2. Find all suspected noisy pixels outside IQR distance
3. Compute the permission distance by two thresholds $T_1$ and $T_2$
4. Return all pixels within $T_1$ and $T_2$ to the nonnoisy pixels
5. Estimate all noisy pixels greater than $T_1$ and $T_2$ by local averaging

## V. EXPERIMENTAL RESULTS

The IQR filter was tested over ten 8-bit gray scale 512×512 images against median filter, figure (5). The IQR filter was found to perform quite well on images corrupted with large window size, figure (4). The Peak Signal to Noise Ratio (PSNR) [12] was used to measure the dissimilarities between the noisy image and the original image, table (2). Also, figures (6), (7), and (8), show differences in PSNR graphically between IQR and median filter.

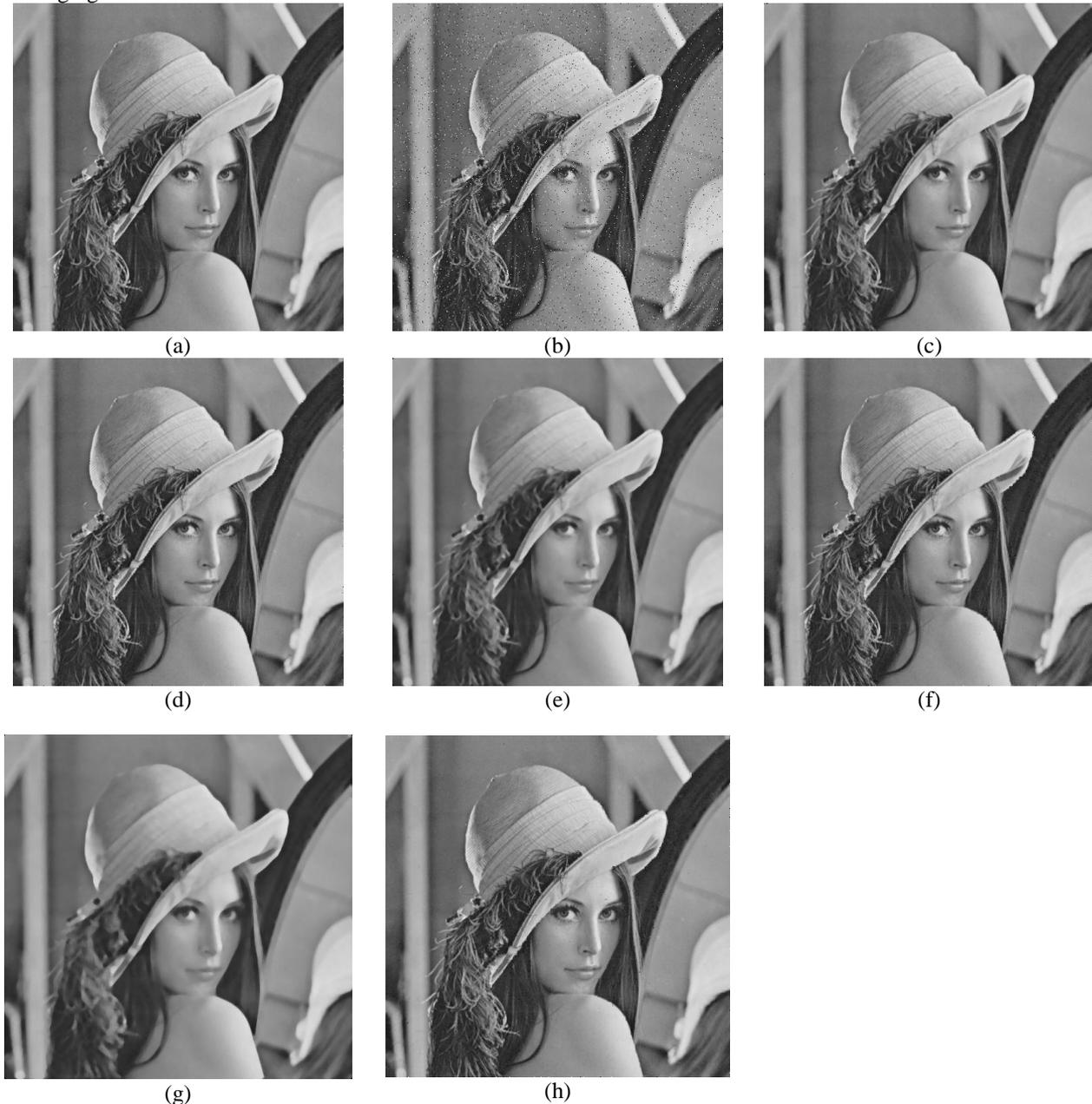

Figure 4 (a) Original Image (b) Noisy Image (c) 3×3 Median Filter (d) 3×3 IQR Filter (e) 5×5 Median Filter (f) 5×5 IQR Filter (g) 7×7 Median Filter (h) 7×7 IQR Filter

TABLE 2 PSNR VALUES FOR TEN 512×512 TEST IMAGES

| # | Image | 3×3 Window Size | | 5×5 Window Size | | 7×7 Window Size | |
|---|---|---|---|---|---|---|---|
| | | Median Filter | IQR filter | Median Filter | IQR filter | Median Filter | IQR filter |
| 1 | Lena | 35.0945 | 38.9235 | 30.9786 | 36.7854 | 28.6724 | 37.3126 |
| 2 | Peppers | 35.8371 | 37.6059 | 32.3491 | 32.9871 | 30.0969 | 32.4500 |
| 3 | Baboon | 22.8738 | 30.2606 | 20.6409 | 31.5668 | 19.9158 | 30.8693 |
| 4 | F16 | 33.9033 | 36.6374 | 29.3291 | 33.6984 | 26.8002 | 33.0441 |
| 5 | Boys | 29.2751 | 31.1613 | 27.5362 | 30.9018 | 26.3659 | 30.6026 |
| 6 | Horse | 26.9739 | 29.8904 | 25.0597 | 29.5346 | 24.5497 | 29.2788 |
| 7 | Lion | 27.3565 | 35.6671 | 25.4204 | 35.1693 | 24.6030 | 34.7828 |
| 8 | Bird | 31.2095 | 33.2074 | 27.7528 | 32.8480 | 25.7590 | 32.5171 |
| 9 | Mosque | 24.2646 | 27.4299 | 22.1538 | 26.5097 | 20.9919 | 26.5175 |
| 10 | Einstein | 35.4273 | 39.3449 | 31.5721 | 39.4829 | 29.8296 | 39.2035 |

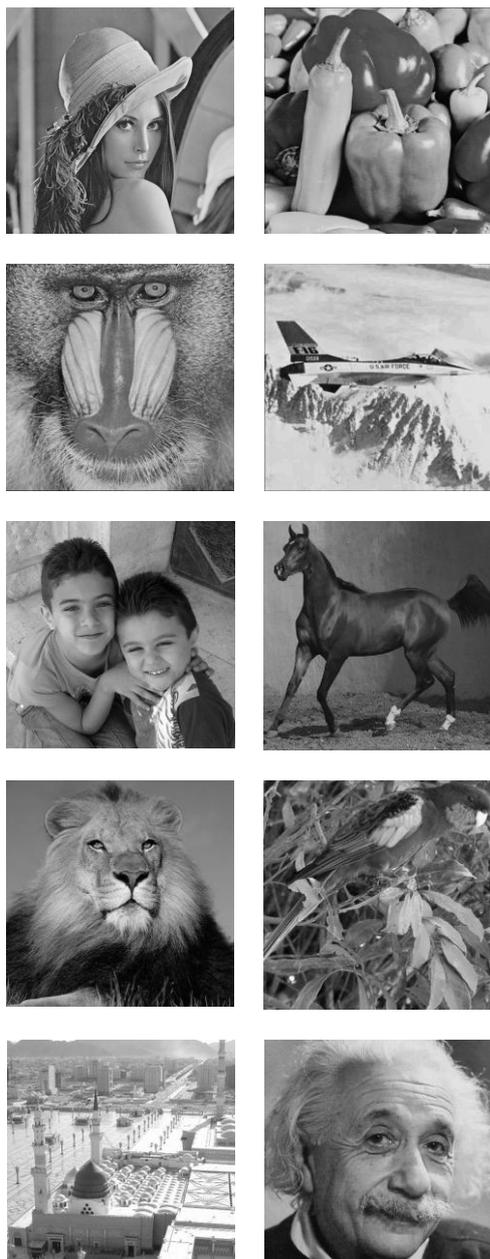

Figure 5 512×512 test images: Lena, peppers, baboon, f16, boys, horse, lion, bird, bird, mosque, and Einstein

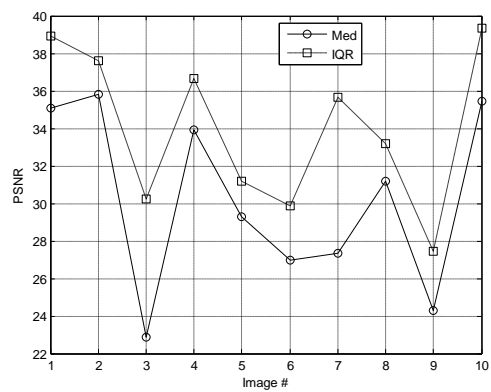

Figure 6 Window of size 3×3

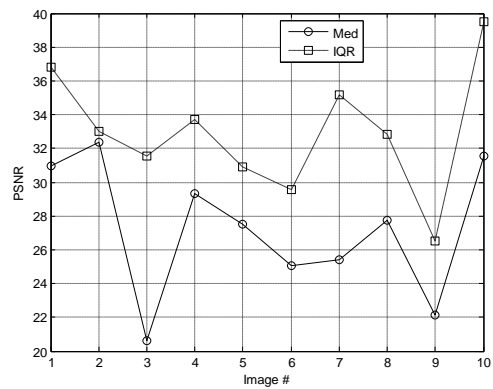

Figure 7 Window of size 5×5

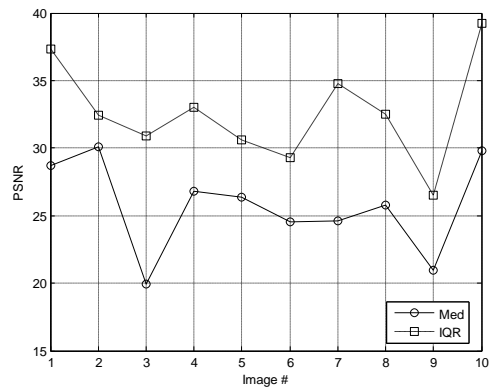

Figure 8 Window of size 7×7

## VI. CONCLUSIONS

In this paper, a new and simple approach for removing salt and pepper noise from corrupted images has been presented. The proposed filter use statistic in a way that removes outlier from a window of size k×k. It can be seen that IQR filter preserves edge sharpness better of the original image than median filter. As a main conclusion from this article is that whenever the window size is increased the preserving of the edges is not affected highly which is on the contrary of the median filter. Results show this filter can effectively reduce salt and pepper noise. However, some problems need to be solved in the future. This algorithm may fail when image regions are spoiled with high noise.

**Firas A. Jassim** received the BS degree in mathematics and computer applications from Al-Nahrain University, Baghdad, Iraq in 1997, and the MS degree in mathematics and computer applications from Al-Nahrain University, Baghdad, Iraq in 1999 and the PhD degree in computer information systems from the university of banking and financial sciences, Amman, Jordan in 2012. His research interests are Image processing, image compression, image enhancement, image interpolation and simulation.